\documentclass[conference]{IEEEtran}
\IEEEoverridecommandlockouts
\usepackage{cite}
\usepackage{amsmath,amssymb,amsfonts}
\usepackage{algorithmic}
\usepackage{subcaption}
\usepackage{graphicx}
\usepackage{textcomp}

\usepackage{xcolor}
\usepackage{array}
\def\BibTeX{{\rm B\kern-.05em{\sc i\kern-.025em b}\kern-.08em
    T\kern-.1667em\lower.7ex\hbox{E}\kern-.125emX}}
\begin{document}

\title{Biologically inspired ChaosNet architecture for Hypothetical Protein Classification\\
}

\author{\IEEEauthorblockN{Sneha K H}
\IEEEauthorblockA{\textit{Department of Mathematics} \\
\textit{Amrita Vishwa Vidyapeetham}\\
Amritapuri, India \\
snehakh@am.students.amrita.edu}
\and
\IEEEauthorblockN{Adhithya Sudeesh}
\IEEEauthorblockA{\textit{Department of Mathematics} \\
\textit{Amrita Vishwa Vidyapeetham}\\
Amritapuri, India \\
adhithyasudeesh@am.students.amrita.edu}
\and
\IEEEauthorblockN{Pramod P Nair}
\IEEEauthorblockA{\textit{Department of Mathematics} \\
\textit{Amrita Vishwa Vidyapeetham}\\
Amritapuri, India \\
pramodpn@am.amrita.edu}
\and
\IEEEauthorblockN{Prashanth Suravajhala}
\IEEEauthorblockA{\textit{Amrita School of Biotechnology} \\
\textit{Amrita Vishwa Vidyapeetham}\\
Amritapuri, India \\
prash@am.amrita.edu}
}

\maketitle

\begin{abstract}
ChaosNet is a type of artificial neural network framework developed for classification problems and is influenced by the chaotic property of the human brain. Each neuron of the ChaosNet architecture is the one-dimensional chaotic map called the Generalized Luröth Series (GLS). The addition of GLS as neurons in ChaosNet makes the computations straightforward while utilizing the advantageous elements of chaos. With substantially less data, ChaosNet has been demonstrated to do difficult classification problems on par with or better than traditional ANNs. In this paper, we use Chaosnet to perform a functional classification of Hypothetical proteins [HP], which is indeed a topic of great interest in bioinformatics. The results obtained with significantly lesser training data are compared with the standard machine learning techniques used in the literature. 
\end{abstract}


\section{Introduction}
Deep Learning algorithms have gathered a lot of popularity among researchers because of their applications in practically all fields of science, engineering, medicine, finance \cite{alligood1998chaos} and agriculture \cite{Nair2011}. Although these algorithms were influenced by neural networks in the human brain, they are not directly connected to the biological functions of the brain, such as learning and memory encoding. The availability of vast quantities of training data is crucial for many algorithms in the AI community. But large volumes of training data might not be available in several real-world situations \cite{Staudemeyer2014}.  ChaosNet is one of the architectures that address this issue \cite{Harikrishnan2019}. 

\subsection{ChaosNet}
The human brain is one of the most sophisticated and poorly understood systems. It is estimated to have about 86 billion neurons \cite{Azevedo2009}, each of which interacts with the others to form a highly complex system. Even though artificial neural networks were modeled after the human brain, they are only slightly connected to how the brain stores and learns memories. Although they have made great strides in difficult tasks like image recognition \cite{Premkumar2022}, natural language processing \cite{Thara2022}, etc they are still far from as efficient as the human mind. To benefit from the human brain's unique potential for learning and to better comprehend the brain, researchers are increasingly concentrating on creating biologically influenced algorithms and architectures \cite{Aihara1990,crook2001novel}.

To utilize some of the greatest features of biological neural networks, ChaosNet was developed in 2019 \cite{Harikrishnan2019}. ChaosNet has been demonstrated to perform difficult classification tasks equally or better than standard ANNs while needing a smaller amount of training data. The individual neurons of ChaosNet are represented by the one-dimensional chaotic map Generalized Luröth Series (GLS). ChaosNet is an intriguing ANN architecture for a variety of applications, including classification problems and memory encoding for retrieval purposes and storage. This is because of the chaotic nature of GLS neurons. Topological transitivity, shannon optimal lossless compression, universal approximation property, and computation of logical operations (AND, XOR, etc.), are some of its beneficial traits \cite{Harikrishnan2019}.

\subsection{Hypothetical Proteins (HPs)}

Proteins are the macromolecules that are present in all living organisms. Since proteins naturally stack up into three-dimensional structures, that are determined by the protein polymer's amino acid sequence, the purpose of a protein is intimately connected to its three-dimensional architecture. Because of this, proteins are remarkably the personification of the change from the one-dimensional world of sequences to the three-dimensional world of molecules with various functions \cite{Thomas2022}. Numerous proteins are predicted to eventually have some function, even though thousands of different proteins have a wide variety of activities. The functional importance of hundreds of genes that encode proteins, particularly those that do not have a function, are not expressed, are distinctive from other genomes, or are frequent, is least understood despite the numerous technologies that have been developed to identify components of cell membrane \cite{Sikosek2014}. The functional relevance of many regions of the genome remaining unknown makes them orphan genes, and the proteins encoding those genes are called hypothetical proteins (HPs). These proteins are very significant because many of them could be linked to diseases in people and, as a result, belong to functional families. Despite not being functionally characterized, they are crucial for acknowledging physiological and biochemical mechanisms. \cite{Nimrod2008}. For example, they can be utilized to identify markers, pharmacological targets, novel structures and functions, as well as early detection and benefits for proteomic and genomic research. \cite{Mohan2012}. 

In this paper, we propose the classification of hypothetical proteins using ChaosNet and compare the results we obtain with the best-known solutions available in the literature. We shall also extend the classification using selected features to obtain the best features for better classification. In the next section of the paper, we present a brief literature review on ChaosNet and hypothetical proteins. In section \ref{III}, we present the methodology adopted for classification. Section \ref{IV} discusses the proposed algorithm's results and analysis, followed by our conclusions in the last section.

\section{Literature review}\label{II}
\subsection{Chaos-Net}

The ability of the brain to display chaos, or the phenomena of complicated random-like, and unstable behavior coming from basic nonlinear deterministic systems, is one of the most intriguing aspects of the brain \cite{alligood1998chaos}. A neural system's sensitivity to little changes in internal functional parameters aids in obtaining the correct response to various inputs. This property of the neural system is similar to the dynamical characteristics of chaotic systems \cite{Babloyantz1996}. It is seen that the brain frequently switches between several states rather than returning to equilibrium after a transitory condition. For this reason, it is proposed that the brain can display diverse behaviors-weak chaos, periodic orbits, and strong chaos for various purposes by changing the functional characteristics of the neurons. As a result, chaotic systems with a broad variety of behaviors enable the brain to quickly adapt to shifting circumstances. The process through which a person encodes, stores, and retrieves information from their environment is referred to as memory in psychology. The information is initially gathered, examined, and then during the encoding stage, the brain translates the language in which the data is stored. The brain's encoding phase of data storage appears to have a chaotic structure. Brain networks with billions of neurons exhibit chaotic behavior, but so do the dynamics of individual neurons at the cellular and subcellular levels \cite{Korn2003}. The brain's ability to transmit and store information is due to the impulse trains the neurons produce. The passage of various ions through the axonal membrane, which results in a change in the membrane's voltage, produces these impulses, also known as action potentials \cite{Aram2017}. Hodgkin and Huxley developed the first dynamical system model for the communication of the axon membrane and ion channels that may generate precise action potentials \cite{Hodgkin1952}. Later, its streamlined variations, including Fitzhugh-Nagumo \cite{fitzhugh1961impulses} and Hindmarsh-Rose model \cite{hindmarsh1984model}, were developed. All these models are chaotic in nature. 

Until 2019, none of the architectures proposed for classification problems exhibited chaos at the tier of single neurons, even though some artificial neural networks (Recurrent Neural Networks) exhibit chaotic dynamics. However, as a theoretical reasoning for the human brain's memory encoding, several chaotic neural architectures were proposed, like the Aihara model developed in 1990 \cite{Aihara1990}. Kozma, Freeman, and their team have developed chaotic models that are based on the mammalian olfactory network to explain how odors are memorized \cite{kozma1999possible}. The functional roles of chaotic neural networks as a dynamic link for long-term memory and short-term memory generators have also been examined by Tsuda et al. \cite{Tsuda1992}. To utilize few of the greatest features of biological neural networks, Harikrishnan and his team developed ChaosNet in 2019 \cite{Harikrishnan2019}. 

\begin{figure}[h]
    \centering
    \subfloat[\centering $T_{skew-tent}$ map ]{{\includegraphics[width=3.95cm]{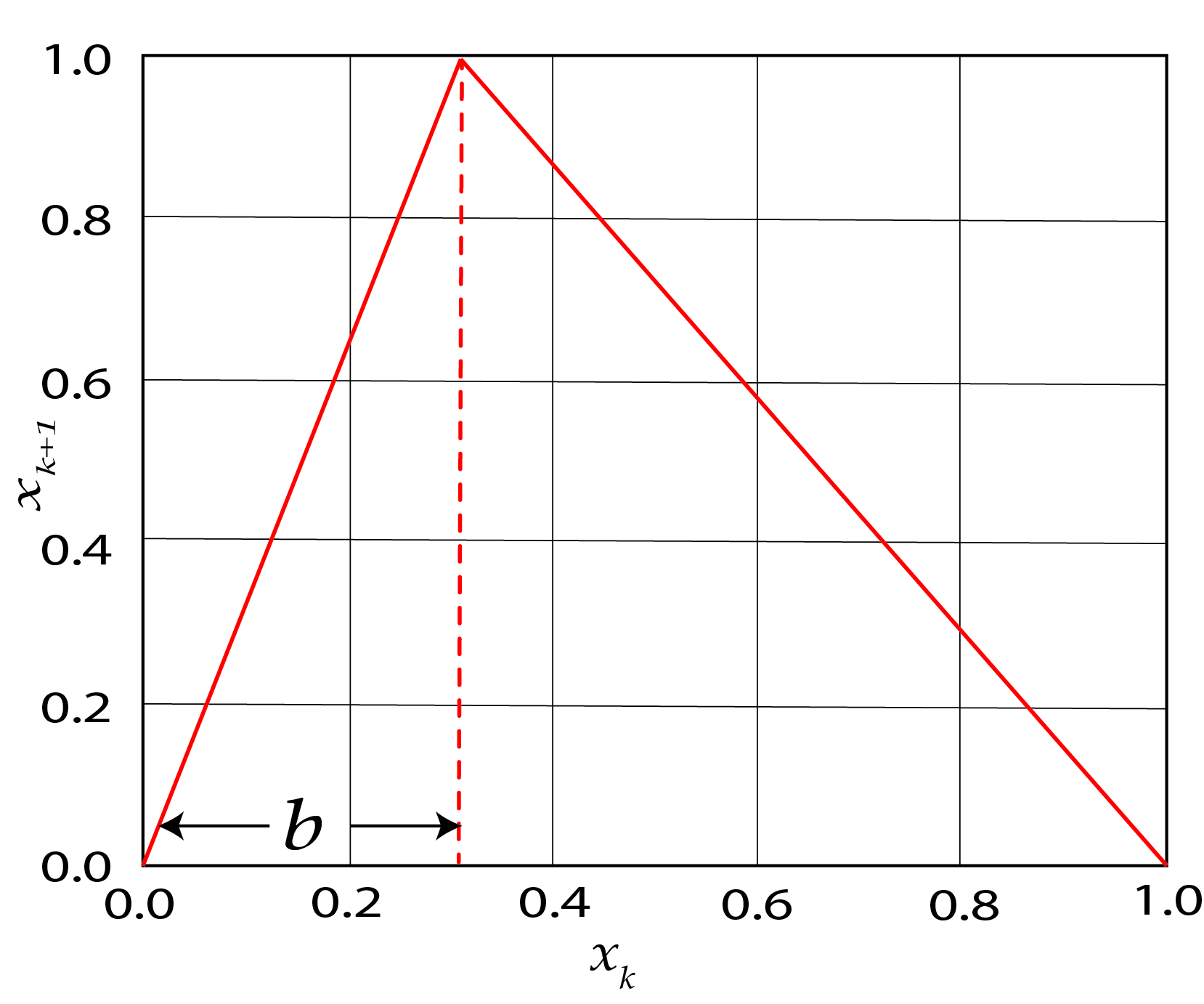} }\label{a}}%
    \qquad
    \subfloat[\centering $T_{skew-binary}$ map]{{\includegraphics[width=3.95cm]{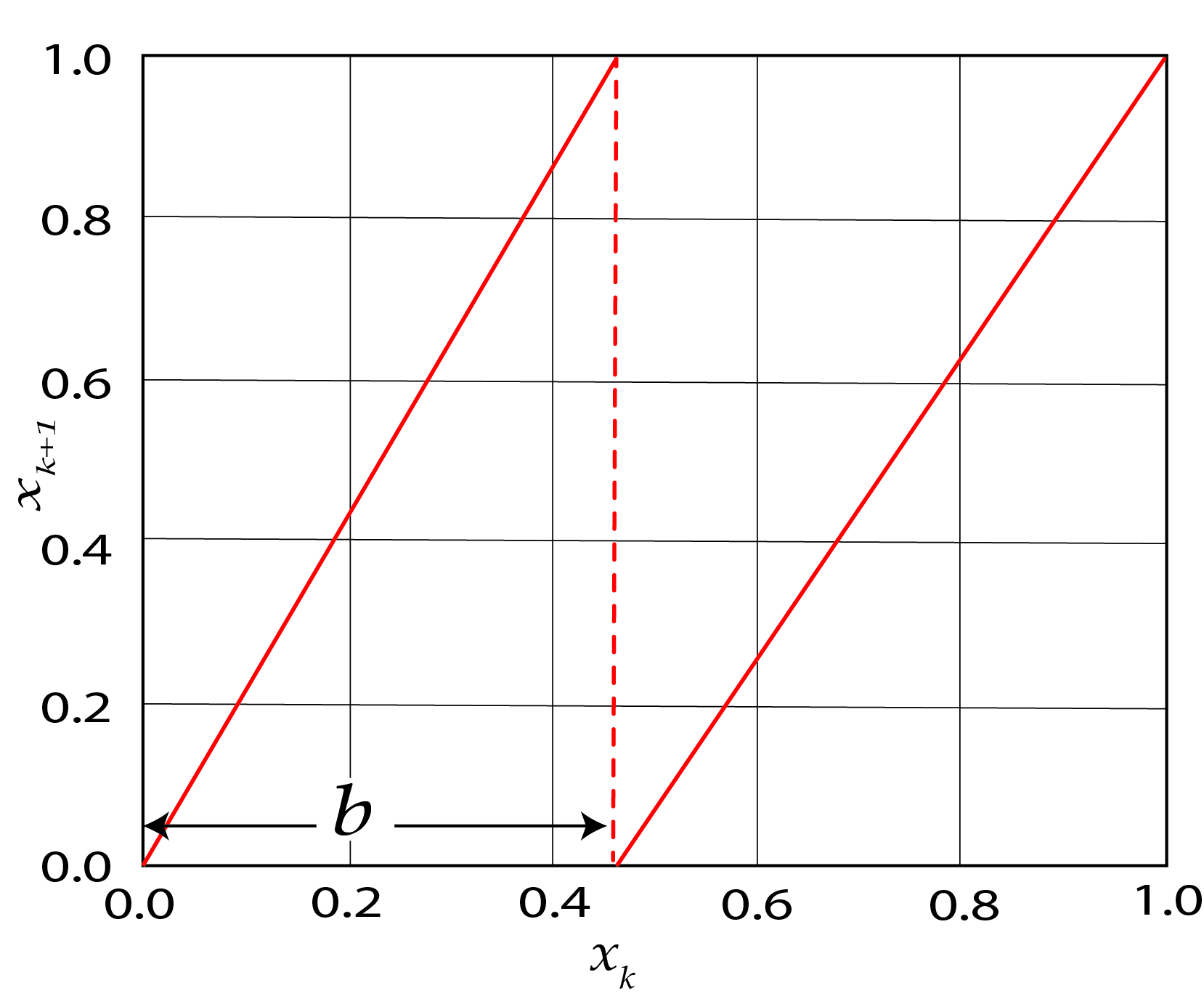} }\label{b}}%
    \caption{Two types of GLS neurons}%
    \label{fig1}%
\end{figure}

ChaosNet has been shown to execute challenging classification tasks as well as or better than standard artificial neural networks while using much less training data. The one-dimensional GLS maps in ChaosNet help keep the processing simple while utilizing the beneficial aspects of chaos. Two types of GLS neurons are used in ChaosNet, namely -  $T_{skew-tent}(x)$ and $T_{skew-binary}(x)$. \\
 $T_{skew-tent}$ : [0,1) $\rightarrow$ [0,1) is defined as:
\begin{equation} \label{eq1}
T_{skew-tent}(x) = 
\begin{cases}
    \begin{array}{lr}
 		x/b, &    0<x<b, \\
		\dfrac{(1-x)}{(1-b)}, & \ b \leq x <1,
    \end{array}
\end{cases}         
\end{equation}
where x $\in$ $[0,1)$  and  $0 < b < 1$. (Fig \ref{a}).\\
$T_{skew-binary}$ : [0,1) $\rightarrow$ [0,1) is defined as:
\begin{equation}
T_{skew-binary}(x) = 
\begin{cases}
    \begin{array}{lr}
 		x/b, &    0<x<b, \\
		\dfrac{(x-b)}{(1-b)}, & \ b \leq x <1,
    \end{array}
\end{cases}         
\end{equation}
where x $\in$ $[0,1)$  and  $0 < b < 1$. (Fig \ref{b}).\\

The technique draws inspiration from biological neurons and leverages the tendency of chaotic neuronal firing as a neuronal code for the learning process. The GLS neuron's property, \textit{Topological Transitivity} is employed for classification purpose here \cite{harikrishnan2019novel}. There are two parameters for each GLS neuron. One is the \textit{internal discrimination threshold}, employed for feature extraction, represented by the parameter $b$. Other is the \textit{initial activity value} $x_0$, which makes the GLS-neuron either fire periodically or chaotically. $b$ regulates the degree of chaos. To combat the impact of noise, GLS-neuron can be enhanced with error-detecting capabilities \cite{Nagaraj2019}. The network can also have a hierarchical structure, which can incorporate details as it is transferred to the network's deeper layers.

\subsection{Annotation of Hypothetical Proteins}
Annotating HPs from a specific genome aids in the discovery of novel structures and functions, which then permits their classification into further protein pathways and cascades. It is crucial to analyze and annotate the roles played by HPs in pathogenic microorganisms that cause various diseases in people and animals since this information will be helpful in docking studies that will support medication discovery. Additionally, the identification of HPs aids in the finding of previously unidentified or previously predicted genes, which is particularly advantageous for genomics \cite{Mohan2012}. To carry out a shared function, proteins frequently interact with one another in a mutually reliant manner. Therefore, based on the interaction partners of proteins, one can deduce their functions \cite{Sivashankari2006}. In addition to predicting biological functions and offering insight into various biochemical processes, the protein-protein interaction (PPI) data offer a strong representation for determining cell organization. Using protein interaction networks (PIN), there has been a recent two-fold rise in PPI data. Numerous sophisticated ways connecting orthology mapping and comparative approaches have been developed to study and visualize proteins. However, the abovementioned bioinformatics approaches do not typically apply to proteins that are ``predicted", ``similar to", or ``hypothetical" but rather to proteins with proven functional relationships \cite{Suravajhala2012}. Genome-wide techniques \cite{Nair2020} have been utilized to clarify the human interactome because high-throughput data are currently scarce for the human proteome. However, one might think of extending the empirically discovered human protein interaction network by utilizing information from protein interaction datasets of the model animals, presuming that functional protein interactions are conserved in evolution. Identifying genes with a common ancestry and, as a result, having the same function between the orthologs was necessary to transfer knowledge from known interaction networks to the unknown. Even after being ``mapped to" or ``linked" with certain known proteins, many known HPs still go unannotated, which makes orthology mapping extremely difficult. Additionally, it can be difficult to analyze large HP datasets because multiple processes must be carried out simultaneously to improve protein annotation \cite{Suravajhala2012}.

In addition to PPIs, there are various approaches to determine whether a protein is necessary, including RNA interference, single-gene deletions, antisense RNA, and transposon mutagenesis. Unfortunately, these techniques are all costly, labor-intensive, and time-consuming. Consequently, to ascertain the function of proteins, computational techniques must be used with high-throughput experimental datasets. \cite{Enright1999}. Based on information gathered from subcellular localization, sequence similarity, homology modeling, phylogenetic profiles, mRNA expression profiles, etc., several computational methods have been developed to ascertain protein function. Long non-coding RNAs \cite{Oyelami2022} have been successfully predicted to localize within the plant cells based on ensemble classifiers using tools like ``IncLocator" and  ``LOCALIZER", which predicts both plant and effector protein localization inside the plant cell's subcellular structure \cite{Cao2018}. On the contrary, combining analyses of all these methods or datasets is thought to be more accurate when incorporating various biological datasets. For the creation of an effective strategy for identifying interactions between proteins, applying characteristics from information theory and machine learning has become increasingly common \cite{You2016}. In their work \cite{Ijaq2019}, Suravajhala and his team took nine significant features of proteins to do their functional annotation. The characteristics that correlate with protein essentiality at the gene level are based on sequencing, topology, and structure. The characteristics are functional linkages, back-to-back orthology, orthology mapping, domain analysis, Homology Modelling, sorting signals and sub-cellular localization, Non-coding RNAs associated with HPs, protein interactions, and Pseudogenes linked to HPs.

\section{Proposed methodology}\label{III}
The hypothetical protein dataset considered for classification is prepared by Suravajhala and team for the nine-point classification scoring schema \cite{Ijaq2019}. The dataset consists of 194 negative instances and 106 positive instances. The positives indicate HPs, while the negatives are functional proteins. We follow a similar methodology proposed by Harikrishnan and his team for classification using \textit{Topological Transitivity - Symbolic Sequence} (TT-SS) based feature extraction. The GLS neuron we use in this work is the skewed tent map defined in equation \ref{eq1}.
\begin{figure} [h]
  \includegraphics[width=\linewidth,scale=0.3]{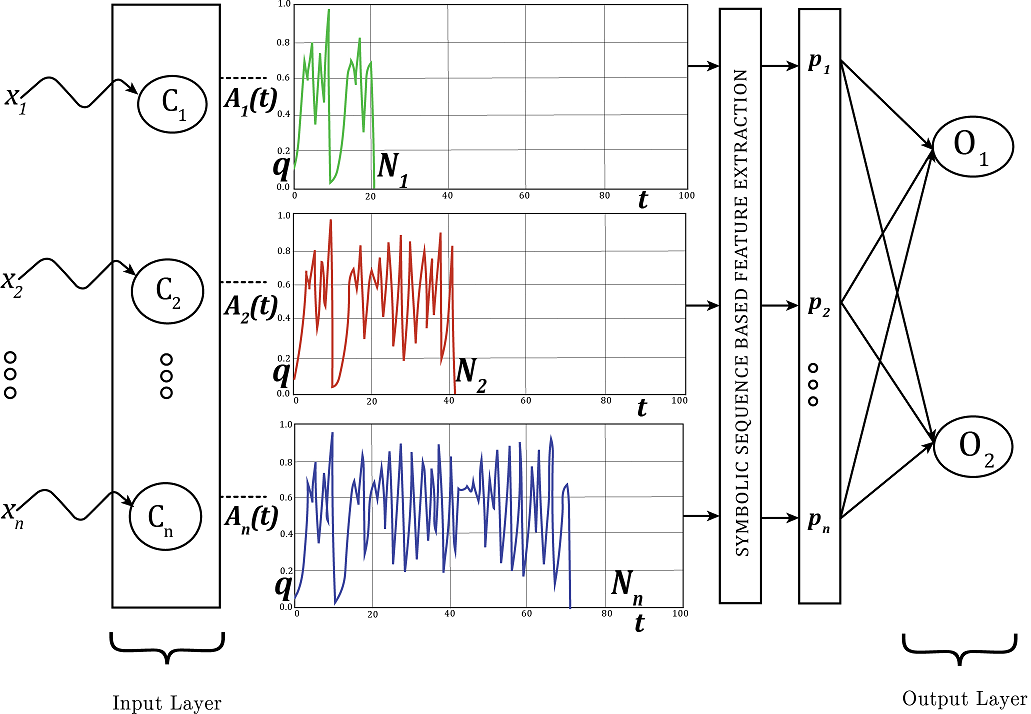}
  \caption{ Architecture of proposed ChaosNet}
  \label{fig2}
\end{figure}

A single-layer chaos-inspired neural network is used for this classification. The $n$ GLS neurons $C_1, C_2,..., C_n$ are composed in the input layer which extracts patterns from input data. The representation vectors for each of the target classes denoted as 1 and 2 are stored in the output layer nodes $O_1$ and $O_2$. An $m\times n$ matrix ($X$) is used to represent the input data, where $m$ denotes the number of data instances and $n$ denotes the number of features. Fig.\ref{fig2} demonstrates the architecture and data flow of ChaosNet. The chaotic map's initial value, or the GLS neurons' initial neural activity, is $q$ units. Upon encountering the stimuli, the GLS neurons begin chaotic firing, which is the input vector of the data instances denoted as $x_1,x_2,...,x_n$. $A_1(t),A_2(t),...,A_n(t)$ are the chaotic neural activity values at time $t$ and is given by
\begin{equation}
A_i(t) \ = \ T_{skew-tent}(A_i(t-1))
\end{equation}

When the activity value $A_i(t)$ corresponding to each GLS neuron starting from $q$ reaches the $\varepsilon$-neighbourhood of $x_i$, their chaotic firing stops. The TT-SS feature ($p_i$) is the portion of this firing time ($N_i$) when the activity value of the GLS neuron is greater than the discriminating threshold ($b$). The Topological Transitivity property for a map $T: R \rightarrow R$ states that for all non-empty open set pairs $D$ and $E$ in $R$, there exists an element $d  \in D$ and a non-negative finite integer $n$ such that $T^{n}(d) \in E$ \cite{Harikrishnan2019}. From this definition, considering $T$ as  GLS-1D map with $R$ as $[0,1)$ and $D$ as  ($q \ - \varepsilon,q + \varepsilon$) and $E$ as  ($x_k \ - \varepsilon$, $x_k + \varepsilon$) where $\varepsilon > 0$. Then there exists an integer $N_k \geq 0$ and a real number $d = q \in D$ such that  $T^{N_k}(d) \in E$ is guaranteed. Here $x_k$ is the stimulus to $k^{th}$ GLS neuron, and $q$ is the initial neural activity. Hence, $N_k \geq 0$ always exists. Firing time $(N_k)$ refers to the amount of time it takes for the firing of the $k^{th}$ GLS neuron to reach the epsilon neighborhood of the incoming stimulus \cite{Harikrishnan2019}.
 
The steps involved in classification are as follows:
\begin{itemize}
\item The dataset is initially normalized and divided into training and testing sets, and the TT-SS feature extraction is done for the training data. 
\item When the neuron is encountered by stimulus $x^{i}_k$, the chaotic firing of the neurons is initiated. The firing trajectory of the GLS neuron starting from $q$ is represented as $q \rightarrow T_k(q) \rightarrow T^{2}_k(q) ... \rightarrow T^{N_i}_k(q)$. 
\item The $\varepsilon$ neighborhood of $x^{i}_k$(stimulus to the kth GLS neuron) is represented as $I^{i}_k = (x^{i}_k - \varepsilon,x^{i}_k + \varepsilon) $ where $\varepsilon > 0$. The chaotic firing halts when the activity value $A(t)$ reaches the $\varepsilon$ neighborhood of $x^{i}_k$.   \item The TT-SS feature is defined as the fraction of firing time where the chaotic trajectory activity value of the GLS-neuron $(A_k(t))$ exceeds the internal discrimination threshold value $(b)$ and is denoted by $p^{i}_k$.
\begin{equation}
p^{i}_k = \frac{h^{i}_k}{N^{i}_k}
\end{equation}
where $h^{i}_k$ is the duration of firing for the kth GLS neuron during which the chaotic trajectory is over the discrimination threshold $(b)$. After the TT-SS feature extraction, we obtain $V^{1}$ and $V^{2}$, corresponding to classes 1 and 2.
\item The average across the rows of $V^{1}$ and $V^{2}$ are taken to obtain the \textit{mean representation vectors} $M^{1}$ and $M^{2}$, corresponding the two target classes. 
\begin{equation}
M^{s} = \frac{1}{m}[\sum_{i=1}^{m} V^{s}_{i1}, \sum_{i=1}^{m} V^{s}_{i2},...,\sum_{i=1}^{m} V^{s}_{in}]
\end{equation}
\item After these representation vectors are stored in the output layer nodes; the TT-SS feature extraction is done for the test data. Let $Z$ be the test data with $i^{th}$ row as $z^i = [z^i_1, z^i_2, z^i_3,..., z^i_n]$.
\item The TT-SS feature extraction is done for test data to obtain $F$ with each row as $f^{i}$.
\item After feature extraction, the cosine similarity of each $f^{i}$ with each of $M^{1}$ and $M^{2}$ is calculated, for which two scalar values will be obtained.
\begin{equation}
\cos(\theta_s )={ {f^{i}} \cdot \ {M^{s}} \over \| {f^{i}} \|_2\| {M^{s}} \|_2}
\end{equation}
 The index ($l$) corresponding to the maximum cosine similarity is the label for $f^{i}$. The process above is repeated until each test data instance has a distinct label.
\end{itemize}
A hyperparameter tuning is done on the hyperparameters of the model - internal discrimination threshold ($b$), initial neural activity ($q$), and $\varepsilon$ that are used to determine the neighborhood interval of $x_{i}$. The best values obtained for the parameters are $b = 0.89, q = 0.499,  \varepsilon = 0.043$.

\section{Results and discussion}\label{IV}
The nine features of proteins used are labeled as follows; 1: Pfam; 2: Orthology; 3:Protein interactions; 4: Best Blast hits; 5: Subcellular localization; 6: Functional linkages; 7: HPs linked to Pseudogenes 8: Homology modeling; 9: HPs linked to ncRNAs \cite{Ijaq2019}. 
We compare the proposed one-layer ChaosNet TT-SS method with the results obtained in \cite{Ijaq2019} using machine learning algorithms like Decision Tree, Naive Bayes, Multi-layer perceptron (MLP), and Support vector machine (SVM).

\begin{table} [h]
\caption{Comparison of accuracies of various algorithms}
	\label{tab: table 1}
\begin{center}
\begin{tabular}{ |c|c|c|  }
\hline
Algorithms & \multicolumn{2}{|c|}{Accuracy ($\%$)}  \\
& 6 point schema & 9 point schema \\
\hline
ChaosNet & 85.33 & 96.34 \\
Naive Bayes & 55.21 & 96.67 \\
SVM & 59.46 & 96.67 \\
Decision Tree  & 48.65 & 96 \\
MLP & 81.08 & 97.67 \\
\hline
\end{tabular} 
\end{center}
\end{table}

A 10-fold cross-validation is done on the dataset before each algorithm is executed. For ChaosNet, we have used only 20\% of the data for training, and the remaining 80\% is used to test the algorithm's accuracy. The other algorithms used for comparison use 66\% for training and the remaining for testing. ChaosNet is also run on the data from the six-point classification schema proposed by Suravajhala et al. in 2012 \cite{Suravajhala2012}. The accuracy obtained by ChaosNet was found to be at par with other algorithms in the nine-point classification schema. At the same time, the results outperformed the other algorithms in the six-point schema. The accuracies obtained are tabulated in table \ref{tab: table 1}.  

The number of instances $m$ taken per class varies from one to twenty. The classification accuracy increases significantly from $80\% (m=1)$ to $96\%(m=20)$. For values of $m$ higher than 20, the algorithm does not give sufficient changes in the classification accuracy. The figure below shows the accuracies for different values of $m$.   
\begin{figure} [h]
\begin{center}
  \includegraphics[width=70mm]{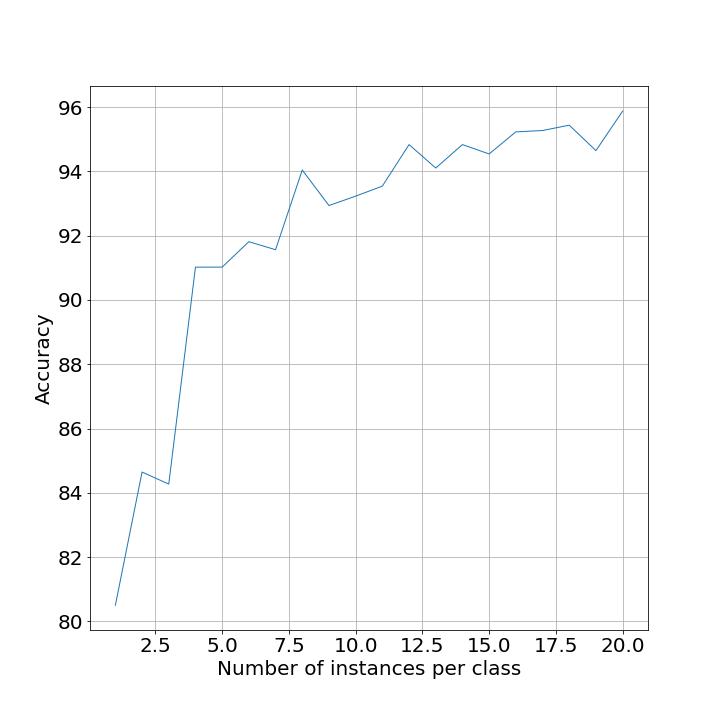}
  \caption{ Classification accuracy for instances per class}
  \label{nn}
\end{center}
\end{figure}

Feature selection was done to obtain features with the best accuracy compared to other algorithms. Features 1,5,7,8,9 gave an accuracy of $96.25\%$ for ChaosNet. The accuracies of other algorithms with the same features are compared and tabulated in table \ref{tab: table 2}.

\begin{table} [h]
	\caption{ Accuracies of algorithms with features 1,5,7,8,9. }
	\label{tab: table 2}
	\begin{center}
		\begin{tabular}{ |c|c| }
			\hline
			Algorithms & Accuracy($\%$) \\
			\hline
			ChaosNet & 96.25 \\
			SVM & 88.33 \\
			Decision Tree  & 95.41 \\
			MLP & 95.41 \\
			Naive Bayes & 95.83 \\
			\hline 
		\end{tabular} 
	\end{center}
\end{table}

\section{Conclusion}\label{V}
A single-layer ChaosNet algorithm is used to classify hypothetical proteins using the six- and nine-point schema. The proposed algorithm used comparatively fewer training samples and a larger test dataset when compared to other machine learning algorithms. Chaosnet outperformed the other algorithms in the data with six features, while the results were equally good as that of other algorithms in the nine features data.

Various other improvements have been proposed on ChaosNet in recent years. Using the multilayer ChaosNet TT-SS method with fully connected layers instead of the single-layer TT-SS method would ensure better accuracies. Another work on chaos introduces a learning algorithm, Neurochaos Learning (NL), which focuses on solving the challenging problem of learning from an imbalanced dataset. Recently, a combination of neurochaos-based feature transformation and extraction with traditional ML algorithms is also proposed. As an extension of this work, we shall compare these newly developed algorithms on the dataset to analyze their performances.

\bibliographystyle{IEEEtran}
\bibliography{chaos}

\begin{thebibliography}{10}
\providecommand{\url}[1]{#1}
\csname url@samestyle\endcsname
\providecommand{\newblock}{\relax}
\providecommand{\bibinfo}[2]{#2}
\providecommand{\BIBentrySTDinterwordspacing}{\spaceskip=0pt\relax}
\providecommand{\BIBentryALTinterwordstretchfactor}{4}
\providecommand{\BIBentryALTinterwordspacing}{\spaceskip=\fontdimen2\font plus
\BIBentryALTinterwordstretchfactor\fontdimen3\font minus
  \fontdimen4\font\relax}
\providecommand{\BIBforeignlanguage}[2]{{%
\expandafter\ifx\csname l@#1\endcsname\relax
\typeout{** WARNING: IEEEtran.bst: No hyphenation pattern has been}%
\typeout{** loaded for the language `#1'. Using the pattern for}%
\typeout{** the default language instead.}%
\else
\language=\csname l@#1\endcsname
\fi
#2}}
\providecommand{\BIBdecl}{\relax}
\BIBdecl

\bibitem{alligood1998chaos}
K.~T. Alligood, T.~D. Sauer, J.~A. Yorke, and D.~Chillingworth, ``Chaos: an
  introduction to dynamical systems,'' \emph{SIAM Review}, vol.~40, no.~3, pp.
  732--732, 1998.

\bibitem{Nair2011}
P.~P. Nair, ``A multigradient algorithm using a mixture of experts architecture
  for land cover classification of multisensor images,'' \emph{International
  journal of remote sensing}, vol.~32, no.~17, pp. 4933--4941, 2011.

\bibitem{Staudemeyer2014}
R.~C. Staudemeyer and C.~W. Omlin, ``Extracting salient features for network
  intrusion detection using machine learning methods,'' \emph{South African
  computer journal}, vol.~52, no.~1, pp. 82--96, 2014.

\bibitem{Harikrishnan2019}
H.~N~B, A.~Kathpalia, S.~Saha, and N.~Nagaraj, ``Chaosnet: A chaos based
  artificial neural network architecture for classification,'' \emph{Chaos: An
  Interdisciplinary Journal of Nonlinear Science}, vol.~29, p. 113125, 11 2019.

\bibitem{Azevedo2009}
F.~A. Azevedo, L.~R. Carvalho, L.~T. Grinberg, J.~M. Farfel, R.~E. Ferretti,
  R.~E. Leite, W.~J. Filho, R.~Lent, and S.~Herculano-Houzel, ``Equal numbers
  of neuronal and nonneuronal cells make the human brain an isometrically
  scaled-up primate brain,'' \emph{Journal of Comparative Neurology}, vol. 513,
  no.~5, pp. 532--541, 2009.

\bibitem{Premkumar2022}
A.~Premkumar, R.~Hridya~Krishna, N.~Chanalya, C.~Meghadev, U.~A. Varma,
  T.~Anjali, and S.~Siji~Rani, ``Sign language recognition: A comparative
  analysis of deep learning models,'' in \emph{Inventive Computation and
  Information Technologies}.\hskip 1em plus 0.5em minus 0.4em\relax Springer,
  2022, pp. 1--13.

\bibitem{Thara2022}
S.~Thara and P.~Poornachandran, ``Social media text analytics of
  malayalam--english code-mixed using deep learning,'' \emph{Journal of big
  Data}, vol.~9, no.~1, pp. 1--25, 2022.

\bibitem{Aihara1990}
K.~Aihara, T.~Takabe, and M.~Toyoda, ``Chaotic neural networks,'' \emph{Physics
  letters A}, vol. 144, no. 6-7, pp. 333--340, 1990.

\bibitem{crook2001novel}
N.~T. Crook and T.~O. Scheper, ``A novel chaotic neural network architecture.''
  in \emph{ESANN}, 2001, pp. 295--300.

\bibitem{Thomas2022}
L.~X. Thomas, P.~Archa, B.~Nair, P.~Suravajhala, and R.~Suravajhala,
  ``Ascertaining structural dynamics for a conformational plasticity in
  lncrna-hsp90 interactions,'' 2022.

\bibitem{Sikosek2014}
T.~Sikosek and H.~S. Chan, ``Biophysics of protein evolution and evolutionary
  protein biophysics,'' \emph{Journal of The Royal Society Interface}, vol.~11,
  no. 100, p. 20140419, 2014.

\bibitem{Nimrod2008}
G.~Nimrod, M.~Schushan, D.~M. Steinberg, and N.~Ben-Tal, ``Detection of
  functionally important regions in “hypothetical proteins” of known
  structure,'' \emph{Structure}, vol.~16, no.~12, pp. 1755--1763, 2008.

\bibitem{Mohan2012}
R.~Mohan and S.~Venugopal, ``Computational structural and functional analysis
  of hypothetical proteins of staphylococcus aureus,'' \emph{Bioinformation},
  vol.~8, no.~15, p. 722, 2012.

\bibitem{Babloyantz1996}
A.~Babloyantz and C.~Louren{\c{c}}o, ``Brain chaos and computation,''
  \emph{International Journal of Neural Systems}, vol.~7, no.~04, pp. 461--471,
  1996.

\bibitem{Korn2003}
H.~Korn and P.~Faure, ``Is there chaos in the brain? ii. experimental evidence
  and related models,'' \emph{Comptes rendus biologies}, vol. 326, no.~9, pp.
  787--840, 2003.

\bibitem{Aram2017}
Z.~Aram, S.~Jafari, J.~Ma, J.~C. Sprott, S.~Zendehrouh, and V.-T. Pham, ``Using
  chaotic artificial neural networks to model memory in the brain,''
  \emph{Communications in Nonlinear Science and Numerical Simulation}, vol.~44,
  pp. 449--459, 2017.

\bibitem{Hodgkin1952}
A.~L. Hodgkin and A.~F. Huxley, ``A quantitative description of membrane
  current and its application to conduction and excitation in nerve,''
  \emph{The Journal of Physiology}, vol. 117, no.~4, pp. 500--544, 1952.

\bibitem{fitzhugh1961impulses}
R.~FitzHugh, ``Impulses and physiological states in theoretical models of nerve
  membrane,'' \emph{Biophysical journal}, vol.~1, no.~6, pp. 445--466, 1961.

\bibitem{hindmarsh1984model}
J.~L. Hindmarsh and R.~Rose, ``A model of neuronal bursting using three coupled
  first order differential equations,'' \emph{Proceedings of the Royal society
  of London. Series B. Biological sciences}, vol. 221, no. 1222, pp. 87--102,
  1984.

\bibitem{kozma1999possible}
R.~Kozma and W.~J. Freeman, ``A possible mechanism for intermittent
  oscillations in the kiii model of dynamic memories-the case study of
  olfaction,'' in \emph{IJCNN'99. International Joint Conference on Neural
  Networks. Proceedings (Cat. No. 99CH36339)}, vol.~1.\hskip 1em plus 0.5em
  minus 0.4em\relax IEEE, 1999, pp. 52--57.

\bibitem{Tsuda1992}
I.~Tsuda, ``Dynamic link of memory—chaotic memory map in nonequilibrium
  neural networks,'' \emph{Neural Networks}, vol.~5, no.~2, pp. 313--326, 1992.

\bibitem{harikrishnan2019novel}
N.~B. Harikrishnan and N.~Nagaraj, ``A novel chaos theory inspired neuronal
  architecture,'' in \emph{2019 Global Conference for Advancement in Technology
  (GCAT)}.\hskip 1em plus 0.5em minus 0.4em\relax IEEE, 2019, pp. 1--6.

\bibitem{Nagaraj2019}
N.~Nagaraj, ``Using cantor sets for error detection,'' \emph{PeerJ Computer
  Science}, vol.~5, p. e171, 2019.

\bibitem{Sivashankari2006}
S.~Sivashankari and P.~Shanmughavel, ``Functional annotation of hypothetical
  proteins--a review,'' \emph{Bioinformation}, vol.~1, no.~8, p. 335, 2006.

\bibitem{Suravajhala2012}
P.~Suravajhala and V.~S. Sundararajan, ``A classification scoring schema to
  validate protein interactors,'' \emph{Bioinformation}, vol.~8, no.~1, p.~34,
  2012.

\bibitem{Nair2020}
P.~P. Nair and R.~Sundaravaradhan, ``An improved upper bound for genome
  rearrangement by prefix transpositions,'' in \emph{2020 Advanced Computing
  and Communication Technologies for High Performance Applications (ACCTHPA)},
  2020, pp. 115--119.

\bibitem{Enright1999}
A.~J. Enright, I.~Iliopoulos, N.~C. Kyrpides, and C.~A. Ouzounis, ``Protein
  interaction maps for complete genomes based on gene fusion events,''
  \emph{Nature}, vol. 402, no. 6757, pp. 86--90, 1999.

\bibitem{Oyelami2022}
F.~O. Oyelami, T.~Usman, P.~Suravajhala, N.~Ali, and D.~N. Do, ``Emerging roles
  of noncoding rnas in bovine mastitis diseases,'' \emph{Pathogens}, vol.~11,
  no.~9, p. 1009, 2022.

\bibitem{Cao2018}
Z.~Cao, X.~Pan, Y.~Yang, Y.~Huang, and H.-B. Shen, ``The lnclocator: a
  subcellular localization predictor for long non-coding rnas based on a
  stacked ensemble classifier,'' \emph{Bioinformatics}, vol.~34, no.~13, pp.
  2185--2194, 2018.

\bibitem{You2016}
Z.-H. You, M.~Zhou, X.~Luo, and S.~Li, ``Highly efficient framework for
  predicting interactions between proteins,'' \emph{IEEE transactions on
  cybernetics}, vol.~47, no.~3, pp. 731--743, 2016.

\bibitem{Ijaq2019}
J.~Ijaq, G.~Malik, A.~Kumar, P.~S. Das, N.~Meena, N.~Bethi, V.~S. Sundararajan,
  and P.~Suravajhala, ``A model to predict the function of hypothetical
  proteins through a nine-point classification scoring schema,'' \emph{BMC
  bioinformatics}, vol.~20, no.~1, pp. 1--8, 2019.

\end{thebibliography}
\end{document}